\PassOptionsToPackage{bookmarks=true}{hyperref}
\PassOptionsToPackage{sort&compress}{natbib}
\documentclass[]{fairmeta}

\usepackage{helvet}

\setcitestyle{numbers,square,comma}

\DeclareTextFontCommand{\textbf}{\fontseries{bx}\selectfont}

\usepackage[T1]{fontenc}
\usepackage[utf8]{inputenc}
\usepackage{latexsym}
\usepackage{inconsolata}
\usepackage{tabularx}
\usepackage{amsmath}
\usepackage{amssymb}
\usepackage{algorithm}
\usepackage{algorithmic}
\usepackage[usestackEOL]{stackengine}
\usepackage[normalem]{ulem}
\usepackage{xspace}
\usepackage{enumitem}
\usepackage[most]{tcolorbox}
\usepackage{wrapfig}
\usepackage{xcolor}
\usepackage{listings}
\usetikzlibrary{calc}
\usepackage{titlesec}
\usepackage{marvosym}
\usepackage{capt-of}

\titleformat{\paragraph}[runin]
  {\normalfont\normalsize\bfseries}
  {\theparagraph}
  {1em}
  {}
  
\setlist[itemize]{
    leftmargin=1em,       
    itemindent=1em,     
    labelsep=0.5em,     
    listparindent=0pt,  
    parsep=0pt          
}

\newcolumntype{Y}{>{\centering\arraybackslash}X}

\tcbset{
  convtitle/.code={\notblank{#1}{\tcbset{title={#1}}}{}},
}
\newtcolorbox{conversationbox}[1][]{%
  enhanced,
  breakable,
  colback=metabg,
  colframe=metablue!55!metabg,
  boxrule=0.8pt,
  arc=4pt,
  outer arc=4pt,
  boxsep=0pt,
  left=8pt,
  right=8pt,
  top=7pt,
  bottom=7pt,
  fontupper=\small\color{metafg},
  parbox=false,
  fonttitle=\small\bfseries,
  coltitle=white,
  colbacktitle=metablue,
  titlerule=0mm,
  toptitle=5pt,
  bottomtitle=5pt,
  lefttitle=10pt,
  righttitle=10pt,
  convtitle={#1},
}

\newcommand{\prismo}{\textit{PRISM}\xspace}

\title{\raisebox{-0.32\height}{\includegraphics[height=1.75em]{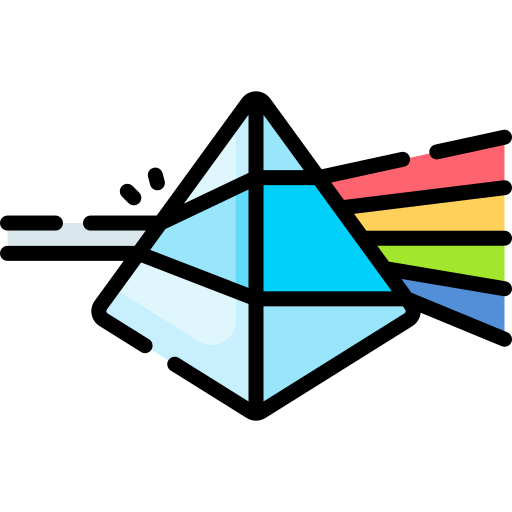}\hspace{-0.3em}}~Don't Mix Rewards, Mix Policies: \\ \vspace{-0.8em}Policy Decomposition and Optimization for \\ \vspace{0.3em}Multi-Reward RL}

\author[1,2,3,*,\ddagger]{Ruiming Liang}
\author[3,4,*,\ddagger]{Yi Zhong}
\author[3]{Yizhen Yuan}  
\author[3]{Yinan Zheng}
\author[3]{Tianyi Tan}
\author[1,2]{\\[0.15em]Tianyue Wang}
\author[1,\text{\Letter}]{Haiyun Guo}

\author[1]{Jinqiao Wang}
\author[3,\text{\Letter}]{Xianyuan Zhan}
\renewcommand\authorformat[2][]{{{\sffamily \bfseries #2$^{#1}$}}}
\makeatletter
\renewcommand\authorformat[2][]{%
  \in@{\\}{#2}%
  \ifin@
    {\sffamily \bfseries #2$^{#1}$}%
  \else
    \mbox{{\sffamily \bfseries #2$^{#1}$}}%
  \fi
}
\makeatother
\renewcommand\affiliationformat[2][]{{\normalsize $^{#1}$#2}}

\affiliation[1]{Fundation Model Research Center, CASIA}
\affiliation[2]{School of Artificial Intelligence, UCAS}
\affiliation[3]{Institute for AI Industry Research (AIR), Tsinghua University}
\affiliation[4]{College of Automotive and Energy Engineering (CAEE), Tongji University}

\contribution[*]{Equal contribution}
\contribution[\text{\Letter}]{Corresponding authors}
\contribution[\ddagger]{Work done during internship at THU-AIR}

\abstract{Modern large language models (LLMs) are expected not just to answer correctly, but to adapt their behavior to different human values and use cases. As a result, multi-reward reinforcement learning (RL) has become an increasingly important problem for LLMs, where each reward captures a different aspect of desired behavior. However, optimizing with multiple rewards suffers from a more severe alignment tax issue, where different optimization objectives can trade off or even conflict with each other, leading to unstable and inefficient post-training.
In this work, we propose \prismo, a new multi-reward RL framework built upon the idea of policy-space decomposition and composition. Instead of compositing different rewards, \prismo optimizes a set of standalone positive policies and a global negative policy. This alleviates the potential conflict during multi-reward policy optimization, while enabling controllability during inference by flexible policy composition.
Experiments on scientific reasoning, tool-use reasoning, and helpfulness–safety alignment show that \prismo consistently outperforms existing multi-reward RL baselines, with extra controllability for inference-time preference control.}

\metadata[Emails]{\texttt{liangruiming2024@ia.ac.cn}, \texttt{zhanxianyuan@air.tsinghua.edu.cn}}

\begin{document}

\newcommand{\zyn}[1]{\textcolor{orange}{\small{\bf [zyn: #1]}}}
\maketitle

\section{Introduction}
\label{sec:intro}

Large language models~\citep{achiam2023gpt, liu2024deepseek} are widely deployed as general-purpose assistants, making alignment with human intent a central concern~\citep{askell2021general, ji2023beavertails}. Reinforcement learning~\citep{sutton1998reinforcement} is among the most effective tools for this purpose, optimizing LLMs directly against reward signals that encode human preferences~\citep{ouyang2022training, bai2022training}. Yet human preferences are rarely captured by a single objective: users expect models to be helpful, safe, and honest at the same time~\citep{wang2023aligning, liu2026gdpogrouprewarddecouplednormalization}, and these goals often conflict, so improving one dimension costs another~\citep{ouyang2022training, yang2024rewards, lin2024mitigating}. LLM alignment is thus inherently a multi-reward optimization problem~\citep{williams2024multi, liu2026gdpogrouprewarddecouplednormalization}, substantially harder than single-objective RL.

Existing multi-reward RL methods differ mainly in how they merge reward signals into one training objective. Linear scalarization~\citep{zhou2024beyond, williams2024multi} sums rewards with manual weights; it is simple but scale-sensitive, and the weights must be retuned per task. Reward shaping and adaptive weighting~\citep{wu2023fine, wang2024interpretable, de2024dynamic, liu2026gdpogrouprewarddecouplednormalization} mitigate scale mismatch, but still collapse all rewards into one scalar before each update, so conflicting preferences compete within the same gradient step. Constrained or multi-objective RL~\citep{achiam2017constrained, roijers2013survey, li2025gradient} treats some rewards as Lagrangian constraints in pursuit of Pareto optimality, but is often hard to optimize and unstable in practice.

Despite their differences, all these methods compose preferences in the \emph{reward space} before policy updates. We argue this principle is fundamentally limiting, for two reasons. First, reward signals differ in scale, distribution, and calibration~\citep{wang2024interpretable, liu2026gdpogrouprewarddecouplednormalization, rame2024warm}, making reward weights an unreliable interface for controlling trade-offs. Second, collapsing multiple preferences into one optimization signal forces them to compete within the same parameter update, yielding a compromise policy that under-optimizes every preference dimension---an effect we call the \emph{multi-reward alignment tax}.

\begin{figure*}[t]
    \centering
    \includegraphics[width=1\linewidth]{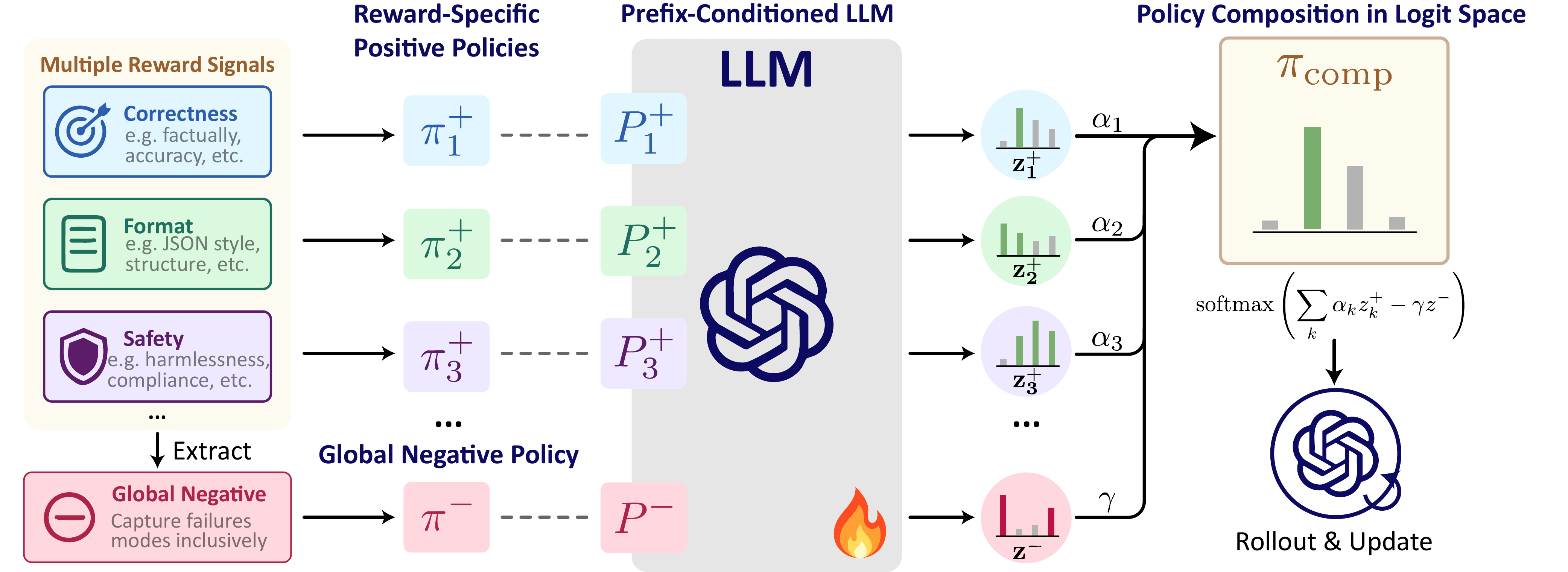}
    \caption{Method overview of \prismo}
    \label{fig:teaser}
    \vspace{-1em}
\end{figure*}

We therefore introduce \prismo, a multi-reward RL algorithm built on a different philosophy: \emph{do not mix rewards, mix policies}. As shown in Figure~\ref{fig:teaser}, \prismo learns one standalone positive policy per reward and a single global negative policy. Desirable behaviors are reward-specific and deserve dedicated optimization directions. In contrast, undesirable behaviors need not be attributed to individual rewards. A single negative policy captures the union of all reward-specific failure modes and penalizes them jointly. Each reward drives updates mainly through its own positive branch. This design alleviates gradient competition among conflicting preferences and mitigates the multi-reward alignment tax.

For efficiency, all sub-policies share a single language model conditioned on different prefix tokens and are composed at sampling time via a weighted sum of logits; the same mixture policy serves both training rollouts and inference. The merge weights thus form an explicit, interpretable interface for preference trade-offs, adjustable at inference time without retraining.

Experiments on three multi-reward settings show that \prismo consistently outperforms reward-space baselines. On GPQA~\citep{rein2023gpqa} and ScienceQA~\citep{saikh2022scienceqa}, it achieves the best overall score on every backbone, surpassing the strongest baseline by 17.8, 8.0, and 0.6 points on DeepSeek-R1-1.5B~\citep{shao2024deepseekmath}, Qwen2.5-1.5B-Instruct, and Qwen2.5-3B-Instruct~\citep{qwen2.5}. On the tool-calling benchmark BFCL-v3~\citep{patil2025bfcl} and the helpfulness--safety benchmarks Alpaca~\citep{taori2023alpaca}, HH-RLHF~\citep{ganguli2022red}, and PKU-SafeRLHF~\citep{dai2024safe}, it ranks first among all RL methods, and remains more robust and controllable as reward complexity grows.

Our contributions are threefold:
\begin{itemize}
    \item We identify the \emph{multi-reward alignment tax} induced by reward-space composition, and advocate composing preferences in policy space instead.
    \item We propose \prismo, which decomposes multi-reward RL into per-reward positive policies and a global negative policy, realized via prefix-conditioned sub-policies with logit-level composition.
    \item Experiments on scientific QA, tool-use reasoning, and helpfulness--safety alignment show consistent gains over strong baselines, with training-free inference-time controllability.
\end{itemize}

\section{Preliminaries}
\label{sec:prelim}

\subsection{Multi-Reward Policy Optimization}
\label{sec:prelim-multireward}

We consider an autoregressive language model parameterized by $\theta$, which defines a conditional distribution $\pi_\theta(o\mid q)$ over responses $o$ given a query $q\sim\mathcal{D}$. Alignment is performed against $N$ reward functions $\{R_k(o,q)\}_{k=1}^{N}$, where each $R_k$ encodes a distinct human preference such as correctness, helpfulness, safety, or format compliance. The native multi-reward objective is the system of $N$ coupled maximization problems

\begin{equation}
\begin{aligned}
        & \mathrm{max}_\pi\ 
        \begin{cases} 
        \mathbb{E}_{o\sim\pi(\cdot\mid q)}[R_1(o,q)], \\
        \mathbb{E}_{o\sim\pi(\cdot\mid q)}[R_2(o,q)], \\
        \cdots\ \\
        \mathbb{E}_{o\sim\pi(\cdot\mid q)}[R_N(o,q)]
        \end{cases}\\
        s.t.\quad & \int_o{\pi(o\mid q)}\mathrm{d}o = 1
\end{aligned}
    \label{eq:multi-reward}
\end{equation}

which, in general, admits no single solution that simultaneously maximizes every $R_k$ \citep{yang2024rewards}. The dominant practical workaround is reward-space scalarization \citep{zhou2024beyond, williams2024multi, liu2026gdpogrouprewarddecouplednormalization}, which collapses these objectives into one weighted scalar $R(o,q)=\sum_{k=1}^{N} w_k R_k(o,q)$ and optimizes a single policy against it. As discussed in Section~\ref{sec:intro}, this composition principle suffers from scale sensitivity, calibration mismatch, and gradient interference between conflicting preferences.

\subsection{Group-Relative Advantage Estimation}
\label{sec:prelim-grpo}

We adopt the group-relative advantage estimator introduced by GRPO~\citep{shao2024deepseekmath} as the building block for our policy updates. Given a query $q$, a group of $G$ responses $\{o_i\}_{i=1}^{G}$ is sampled from the current policy and scored under a reward $R$ to obtain $\{r_i\}_{i=1}^{G}$. The group-relative advantage of response $i$ is
\begin{equation}
    \hat{A}_i \;=\; \frac{r_i - \mathrm{mean}(\{r_j\}_{j=1}^{G})}{\mathrm{std}(\{r_j\}_{j=1}^{G})},
    \label{eq:grpo-adv}
\end{equation}
which provides a low-variance, group-normalized signal without an explicit value network. 

Given the group-relative advantage $\hat{A}_i$ in Eq.~\ref{eq:grpo-adv} computed under a reward function $R$, the corresponding GRPO training objective for a policy $\pi_\theta$ takes the standard clipped surrogate form:
\begin{equation}
\mathcal{L}(\theta) \;=\; -\, \mathbb{E}_{q \sim \mathcal{D},\, \{o_i\}_{i=1}^{G} \sim \pi_{\theta_{\text{old}}}}
\left[ \min\Big( r_i(\theta)\, \hat{A}_i,\ \mathrm{clip}\big(r_i(\theta),\, 1{-}\epsilon,\, 1{+}\epsilon\big)\, \hat{A}_i \Big) \right],
\label{eq:grpo_loss}
\end{equation} 

where $r_i(\theta) = \pi_\theta(o_i \mid q) / \pi_{\theta_{\text{old}}}(o_i \mid q)$ is the importance ratio and $\epsilon$ is the clipping threshold. We write $\mathcal{L}(\theta; R)$ as an explicit function of the scoring reward $R$.

In the multi-reward setting, GDPO \citep{liu2026gdpogrouprewarddecouplednormalization} mitigates reward-scale mismatch by normalizing each reward-specific advantage within the response group, then summing and re-normalizing the aggregated advantage to preserve relative differences across reward combinations.
\begin{equation}
\begin{aligned}
    A_i^k&=\frac{r_i^k - \mathrm{mean}(\{r_j^k\}_{j=1}^{G})}{\mathrm{std}(\{r_j^k\}_{j=1}^{G})}, \quad
    A_i^\mathrm{sum}=\sum_{k=1}^NA_i^k,\\
    \hat{A}_i&=\frac{A_i^\mathrm{sum}-\mathrm{mean}(\{A_j^\mathrm{sum}\}_{j=1}^{G})}{\mathrm{std}(\{A_j^\mathrm{sum}\}_{j=1}^{G})},
\end{aligned}
\end{equation}
where $G$ is group size and $\hat{A}_i$ is GDPO advantage for $i$-th response. Nevertheless, this normalization still amounts to composing rewards in advantage space before policy updates, and thus remains a form of implicit reward weighting that is susceptible to the same gradient interference among conflicting preferences.

\subsection{Positive--Negative Policy Decomposition}
\label{sec:prelim-dual}

Following DIPOLE \citep{liang2025dichotomous} and DExperts \citep{liu2021dexperts}, the optimal policy $\pi^\star$ under a reward function $R(o,q)$ can be expressed as a \textit{product of expert} (PoE) \citep{hinton2002training} composition of a positive policy and a negative policy:
\begin{equation}
\begin{aligned}
    &\pi^\star(o\mid q)=\;\pi^+(o\mid q)^\alpha\cdot\pi^-(o\mid q)^{1-\alpha},
\end{aligned}
\label{eq:pos-neg-decomp}
\end{equation}
where $\pi^{+}$ is the positive policy captures behaviors encouraged by the reward, whereas $\pi^{-}$ is the negative policy captures behaviors suppressed by the reward. The coefficient $\alpha$ controls the degree of greediness in the composed policy; increasing $\alpha$  encourages the final policy to assign higher probability to responses associated with larger rewards.

\section{Method}

\subsection{Multi-Reward Positive-Negative Policy Optimization}

\paragraph{Multi-Reward Policy Optimization}

Consider a multi-objective optimization problem with $N$ rewards $\{R_k(o,q)\}_{k=1}^N$ as in Eq.~\ref{eq:multi-reward}. We aim to train once and allow the trade-off to be freely selected at inference time. To achieve it, assuming that there is an \emph{arbitrary} set of coefficients $\{\alpha_1, \dots, \alpha_N\}$, for any such choice, we formulate a KL-regularized one-step local improvement over a reference policy $\mu$~\citep{ziegler2019fine}, augmented with a shared penalty term:

\begin{equation}
\begin{aligned}
        & \max\limits_{\pi}\ 
        \mathbb{E}_{o\sim\pi(\cdot\mid q)}
        \left[
        \sum_{k=1}^{N}\alpha_k R_k(o,q)
        \right]
        -\frac{1}{\eta}D_{\mathrm{KL}}(\pi\|\mu)
        \\
        &\qquad
        -\gamma
        \mathbb{E}_{o\sim\pi(\cdot\mid q)}
        [
        \psi_{[R_1,R_2,\cdots,R_N]}(o,q)
        ]
        \\
        \mathrm{s.t.}\quad&
        \int_o\pi(o\mid q)\mathrm{d}o=1,
        \\
        &
        \sum_{k=1}^{N}\alpha_k-\gamma=1 .
\end{aligned}
\label{eq:ipv}
\end{equation}

Here $\psi_{[R_1,\dots,R_N]}(o,q)$, abbreviated as $\psi(\mathbf{R})$ below, is designed to model the \textbf{union} of all reward-specific failure modes: it outputs a large value whenever \emph{any one} of the individual rewards is low, regardless of how the others behave, so that a single severely violated objective is sufficient to trigger the penalty. Mathematically, $\psi$ is monotone \textbf{non-increasing} in each reward.

\paragraph{Optimal Policy and Policy Decomposition}

The optimal solution $\pi^\star(o\mid q)$ of Eq.~\ref{eq:ipv} satisfies~\citep{rafailov2023direct} $\pi^\star(o\mid q)=\frac{1}{Z(q)}\mu(o\mid q)\exp\left(\eta\left[\sum_{k=1}^{N}\alpha_kR_k(o,q)-\gamma\psi(\mathbf{R})(o,q)\right]\right)$. Since the constraint in Eq.~\ref{eq:ipv} requires $\sum_{k=1}^N \alpha_k - \gamma = 1$, we can decompose $\mu$ as $\mu(o \mid q) \;=\; \frac{\prod_{k=1}^{N} \mu(o \mid q)^{\alpha_k}}{\mu(o \mid q)^{\gamma}}$. Substituting it into the expression for $\pi^\star$, we obtain

\begin{equation}
\pi^\star(o \mid q) \;=\; \frac{1}{Z(q)} \cdot \frac{\prod_{k=1}^{N} \mu(o \mid q)^{\alpha_k}}{\mu(o \mid q)^{\gamma}} \cdot \frac{\prod_{k=1}^{N} \exp\big(\alpha_k \eta R_k(o,q)\big) } {\exp\big(\gamma \eta \psi(\mathbf{R})(o,q)\big)},
\end{equation}

which can be regrouped term-by-term as
\begin{equation}
\pi^\star(o \mid q) \;=\; \frac{1}{Z(q)} \cdot \frac{\prod_{k=1}^{N} \big[\mu(o \mid q) \exp(\eta R_k(o,q))\big]^{\alpha_k}}{\big[\mu(o \mid q) \exp(\eta \psi(\mathbf{R})(o,q))\big]^{\gamma}}.
\label{eq:pistar_regrouped}
\end{equation}

Note that each bracketed term $\mu(o\mid q)\exp(\eta R_k(o,q))$ and $\mu(o\mid q)\exp(\eta \psi(\mathbf{R})(o,q))$ is, up to normalization, exactly the optimal solution of a single-reward KL-regularized optimal policy.

We define the reward-specific positive policies
\begin{equation}
\pi_k^+(o \mid q) \;\triangleq\; \frac{1}{Z_k^+(q)}\, \mu(o \mid q) \exp\big(\eta R_k(o,q)\big), \qquad k = 1, \dots, N, 
\label{eq:pos_def}
\end{equation}
and the global negative policy
\begin{equation}
\begin{aligned}
\pi^-(o \mid q) \;\triangleq\; \frac{1}{Z^-(q)}\, \mu(o \mid q) \exp\big(\eta \psi(\mathbf{R})(o,q)\big), \\
\end{aligned}
\label{eq:neg_def}
\end{equation}
where $r^+_k=\pi^+_k(o\mid q)/\mu(o\mid q)$, $r^-=\pi^-(o\mid q)/\mu(o\mid q)$, and $Z_k^+(q)$ and $Z^-(q)$ are the corresponding normalizing constants, then $\pi^\star$ can be expressed as a composition of the positive and negative policies~\citep{hinton2002training}:
\begin{equation}
\pi^\star(o \mid q) \;\propto\; \frac{\prod_{k=1}^{N} \big(\pi_k^+(o \mid q)\big)^{\alpha_k}}{\big(\pi^-(o \mid q)\big)^{\gamma}}.
\label{eq:pistar_decomp}
\end{equation}

The derivation of Eq.~\ref{eq:pistar_decomp} shows that, at the optimum, policy composition is decoupled from the optimization of each sub-policy. In particular, the definitions of $\pi_k^+$ and $\pi^-$ do not involve the coefficients $\{\alpha_k, \gamma\}$ used to compose them at inference time. We therefore optimize each sub-policy on its own with the GRPO loss defined in Eq.~\ref{eq:grpo_loss}:

\begin{equation}
\mathcal{L}_k^+(\pi^+_k) \;=\; -\, \mathbb{E}_{q \sim \mathcal{D},\, \{o_i\}_{i=1}^{G} \sim \mu}
\left[ \min\Big( r_i^k\, \tilde{A}_i^k,\ \mathrm{clip}\big(r_i^k,\, 1{-}\epsilon,\, 1{+}\epsilon\big)\, \tilde{A}_i^k \Big) \right],
\label{eq:pos_loss}
\end{equation}
\begin{equation}
\mathcal{L}^-(\pi^-) \;=\; -\, \mathbb{E}_{q \sim \mathcal{D},\, \{o_i\}_{i=1}^{G} \sim \mu}
\left[ \min\Big( r_i^- \tilde{A}_i^-,\ \mathrm{clip}\big(r_i^-,\, 1{-}\epsilon,\, 1{+}\epsilon\big)\, \tilde{A}_i^- \Big) \right].
\label{eq:neg_loss}
\end{equation}

In practice, Eq.~\ref{eq:pos_loss} and Eq.~\ref{eq:neg_loss} use monotonically shaped advantages
$\tilde{A}_i^k \triangleq \sigma(\hat{A}_i^k)-c,\quad
\tilde{A}_i^- \triangleq c-\Big(\prod_{k=1}^{N}\sigma(\hat{A}_i^k)\Big)^{1/N}$
where $\sigma$ is the sigmoid function and $c$ is a constant offset. The soft conjunction makes $\tilde{A}_i^-$ large on the union of failure modes, as required of $\psi$ (alternatives are ablated in Appendix~\ref{sec:ablation_neg_weight}). Moreover, training proceeds iteratively in practice, with the current composed policy $\pi^\star$ serving as the reference $\mu$ that generates rollouts and anchors the importance ratios at each iteration.

\paragraph{Logit-Level Policy Composition}

Since $\pi^\star$, $\pi_k^+$, and $\pi^-$ share the same reference policy $\mu$ and the same context, taking logarithms converts the product and quotient into a linear combination. The log-partition terms $\log Z(q)$, $\log Z_k^+(q)$, and $\log Z^-(q)$ collapse into a single token-independent constant. This constant is automatically absorbed by the softmax normalization and therefore does not need to be explicitly computed.

Consequently, composing the $N+1$ sub-policies reduces to a weighted summation at the token-logit level~\citep{o2023contrastive, dekoninck2023controlled}:

\begin{equation}
z_t^\star
=
\sum_{k=1}^{N}
\alpha_k z_{k,t}^{+}
-
\gamma z_t^- ,
\label{eq:logit_composition}
\end{equation}

where $z_t^\star$, $z_{k,t}^{+}$, and $z_t^-$ denote the logits of $\pi^\star(\cdot\mid q,o_{<t})$, $\pi_k^+(\cdot\mid q,o_{<t})$, and $\pi^-(\cdot\mid q,o_{<t})$, respectively.

\begin{conversationbox}
\paragraph{Discussion} By decoupling policy optimization from reward trade-off selection, PRISM allows a single training run to yield an entire family of policies spanning all possible coefficient combinations, rather than a single policy tied to one fixed trade-off. This design simultaneously enables inference-time controllability and reduces gradient interference across conflicting objectives, alleviating the multi-reward alignment tax.
\end{conversationbox}

\subsection{Efficient Implementation}
\label{sec:method-practical}


Naively instantiating each sub-policy as an independent model would multiply both parameter count and decoding latency by $N+1$. We avoid this via two choices: (i) all sub-policies share a single backbone conditioned on lightweight, independently-trainable prefixes, and (ii) rollouts under $\pi^\star$ are generated in batched forwards pass rather than $N+1$ sequential ones. Asymmetric update scheme shields the shared backbone from negative branch, making them compatible.

\paragraph{Prefix-conditioned shared backbone.} We realize all sub-policies using one shared language model $\pi_\theta$, conditioned on a set of learnable continuous prefix embeddings $\mathcal{P}=\{P_1^+,P_2^+,\dots,P_{N}^+,P^-\}$~\citep{li2021prefix, lester2021power}, one prefix per sub-policy:

\begin{equation}
    \pi_k^{+}(\cdot\mid q) \triangleq \pi_\theta\!\big(\cdot\,\big|\,[q,P_k^+]\big), \quad
    \pi^{-}(\cdot\mid q) \triangleq \pi_\theta\!\big(\cdot\,\big|\,[q,P^-]\big), \quad
    k=1,\dots,N.
\label{eq:prefix-policy}
\end{equation}

\paragraph{Backbone-preserving asymmetric training.}
The positive branches $\pi_k^+$ are trained on their respective reward signals, whereas $\pi^-$ (Eq.~\ref{eq:neg_def}) is trained to \emph{maximize} $\psi(\mathbf{R})$, thereby capturing the \emph{union} of reward-specific failure modes, i.e., responses that fail on any one objective. To keep the shared backbone aligned with the positive behaviors, we adopt an asymmetric update scheme that updates $\theta$ only through the positive branches~\citep{zhang2024negative}. Let $\delta$ be the learning rate:
\begin{align}
\theta &\leftarrow \theta - \delta\, \nabla_{\theta}\, \mathcal{L}^+_k, &P_k &\leftarrow P_k - \delta\, \nabla_{P^+_k}\, \mathcal{L}^+_k, \label{eq:positive_update} \\
\theta &\leftarrow \theta, &P^-  &\leftarrow P^- - \delta\, \nabla_{P^-}\, \mathcal{L}^-. \label{eq:negative_update}
\end{align}
Positive updates (Eq.~\ref{eq:positive_update}) backpropagate through both $\theta$ and $P_k^+$. Negative updates (Eq.~\ref{eq:negative_update}) apply a stop-gradient to $\theta$, so only $P^-$ is updated. Consequently, the backbone is trained exclusively on high-quality data, while $P^-$ still learns a catch-all representation of the union of failure modes that can suppress them under Eq.~\ref{eq:logit_composition} at inference time.

\begin{figure}[t]
    \centering
    \includegraphics[width=0.8\linewidth]{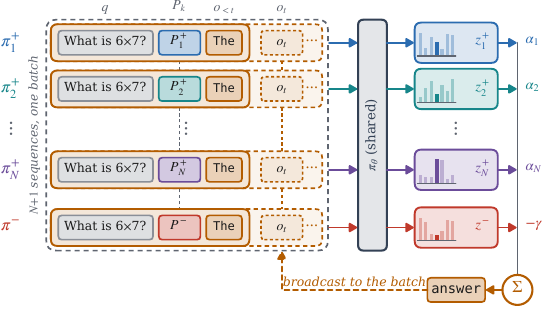}
    \caption{Parallel-batch mixture sampling. At each decoding step, the query $q$, per-branch prefix ($P_k^+$ or $P^-$), and the shared context $o_{<t}$ are stacked along the batch dimension and processed by the shared backbone $\pi_\theta$ in a single forward pass. The resulting per-branch logits $z_k^+$ and $z^-$ are combined with weights $\alpha_k$ and $-\gamma$ (Eq.~\ref{eq:logit_composition}), one token is sampled from $\pi^\star$, and it is broadcast back to all branches for the next step.}
    \label{fig:parallel_batch_sampling}
\end{figure}

\paragraph{Parallel-batch mixture sampling.} Even with a shared backbone, naively sampling from $\pi^\star$ would decode the $N{+}1$ branches sequentially at every step. Instead, we place the branches along the \emph{batch} dimension: as shown in Figure~\ref{fig:parallel_batch_sampling}, the decoding state $(q, o_{<t})$ is replicated across all $N{+}1$ prefixes, their logits are computed in a \textbf{single batched forward pass} and composed via Eq.~\ref{eq:logit_composition}, and the sampled token $o_t$ is broadcast back to all branches so that their KV caches advance in lockstep. Memory and FLOPs still scale linearly with $N{+}1$, but per-token latency is that of a \emph{single} forward pass~\citep{pope2022efficiently}. The same procedure generates rollouts during training and serves inference.

\section{Experiments}

\newcolumntype{Y}{>{\centering\arraybackslash}X}

\begin{table*}[htbp]
\centering
\small
\setlength{\tabcolsep}{2pt}
\renewcommand{\arraystretch}{1.1}

\newcommand{\base}[1]{\textcolor{gray}{#1}}
\newcommand{\gain}[2]{\mbox{#1\textsubscript{\textnormal{\textcolor{gray}{#2}}}}}
\newcommand{\topv}[1]{\textbf{#1}}
\vspace{-1em}
\caption{
\textbf{Main results on scientific QA benchmarks}: scientific QA results across all evaluated backbones, where small gray values denote absolute changes over the corresponding base model. All numbers are percentages. Rows with ``+'' denote optimization methods applied to the corresponding backbone. Bold indicates the best result among optimization methods. Fmt: Percentage of rewards that meet the format requirements. Acc: Percentage of correct answers. Joint: Percentage of answers that meet the format requirements and are correct. Avg: Average of format and accuracy.
}
\label{tab:main_results}
\vspace{-1em}
\resizebox{\ifdim\width>\textwidth\textwidth\else\width\fi}{!}{%
\begin{tabular}{l*{9}{c}}
\toprule
\multicolumn{10}{c}{\textbf{Results on Scientific QA \& GPQA}} \\
\midrule
\multirow{2}{*}{Model / Method}
& \multicolumn{4}{c}{ScienceQA}
& \multicolumn{4}{c}{GPQA}
& \multirow{2}{*}{Overall} \\
\cmidrule(lr){2-5} \cmidrule(lr){6-9}
& Fmt & Acc & Joint & Avg
& Fmt & Acc & Joint & Avg
&  \\
\midrule

\base{DeepSeek-R1-1.5B}
& \base{0.00} & \base{66.64} & \base{0.00} & \base{22.21}
& \base{0.00} & \base{29.68} & \base{0.00} & \base{9.89}
& \base{16.05} \\
\quad + GRPO Sum
& \gain{74.11}{+74.11} & \gain{67.85}{+1.21} & \gain{54.02}{+54.02} & \gain{65.33}{+43.12}
& \gain{16.96}{+16.96} & \gain{29.46}{-0.22} & \gain{6.25}{+6.25} & \gain{17.56}{+7.67}
& \gain{41.44}{+25.39} \\
\quad + GRPO Product
& \gain{74.39}{+74.39} & \gain{68.51}{+1.87} & \gain{55.42}{+55.42} & \gain{66.11}{+43.90}
& \gain{27.68}{+27.68} & \gain{30.80}{+1.12} & \gain{9.60}{+9.60} & \gain{22.69}{+12.80}
& \gain{44.40}{+28.35} \\
\quad + GDPO
& \gain{80.37}{+80.37} & \gain{67.94}{+1.30} & \gain{58.04}{+58.04} & \gain{68.78}{+46.57}
& \gain{26.34}{+26.34} & \gain{28.35}{-1.33} & \gain{8.71}{+8.71} & \gain{21.13}{+11.24}
& \gain{44.96}{+28.91} \\
\quad + \prismo (ours)
& \topv{\gain{95.33}{+95.33}} & \topv{\gain{69.91}{+3.27}} & \topv{\gain{68.51}{+68.51}} & \topv{\gain{77.92}{+55.71}}
& \topv{\gain{81.25}{+81.25}} & \topv{\gain{33.04}{+3.36}} & \topv{\gain{28.35}{+28.35}} & \topv{\gain{47.55}{+37.66}}
& \topv{\gain{62.73}{+46.68}} \\

\midrule

\base{Qwen2.5-1.5B-Instruct}
& \base{6.82} & \base{73.36} & \base{5.89} & \base{28.69}
& \base{37.28} & \base{27.23} & \base{10.49} & \base{25.00}
& \base{26.84} \\
\quad + GRPO Sum
& \topv{\gain{100.00}{+93.18}} & \gain{76.26}{+2.90} & \gain{76.26}{+70.37} & \gain{84.17}{+55.48}
& \gain{77.46}{+40.18} & \gain{27.90}{+0.67} & \gain{21.21}{+10.72} & \gain{42.19}{+17.19}
& \gain{63.18}{+36.34} \\
\quad + GRPO Product
& \gain{98.31}{+91.49} & \gain{75.98}{+2.62} & \gain{74.30}{+68.41} & \gain{82.86}{+54.17}
& \gain{37.95}{+0.67} & \gain{27.90}{+0.67} & \gain{11.16}{+0.67} & \gain{25.67}{+0.67}
& \gain{54.27}{+27.43} \\
\quad + GDPO
& \gain{99.93}{+93.11} & \gain{76.64}{+3.28} & \gain{76.26}{+70.37} & \topv{\gain{84.28}{+55.59}}
& \gain{76.56}{+39.28} & \gain{29.01}{+1.78} & \gain{21.43}{+10.94} & \gain{42.33}{+17.34}
& \gain{63.30}{+36.46} \\
\quad + \prismo (ours)
& \gain{99.53}{+92.71} & \topv{\gain{76.72}{+3.36}} & \topv{\gain{76.45}{+70.56}} & \gain{84.23}{+55.54}
& \topv{\gain{87.95}{+50.67}} & \topv{\gain{29.24}{+2.01}} & \topv{\gain{25.67}{+15.18}} & \topv{\gain{58.45}{+33.45}}
& \topv{\gain{71.34}{+44.50}}
\\

\midrule

\base{Qwen2.5-3B-Instruct}
& \base{44.67} & \base{81.78} & \base{36.73} & \base{54.39}
& \base{47.99} & \base{33.93} & \base{16.52} & \base{32.81}
& \base{43.60} \\
\quad + GRPO Sum
& \topv{\gain{99.81}{+55.14}} & \gain{83.55}{+1.77} & \gain{83.55}{+46.82} & \gain{88.97}{+34.58}
& \gain{87.95}{+39.96} & \gain{29.91}{-4.02} & \gain{27.46}{+10.94} & \gain{48.44}{+15.63}
& \gain{68.71}{+25.11} \\
\quad + GRPO Product
& \gain{99.53}{+54.86} & \gain{84.02}{+2.24} & \gain{84.02}{+47.29} & \topv{\gain{89.19}{+34.80}}
& \gain{84.82}{+36.83} & \gain{28.80}{-5.13} & \gain{24.11}{+7.59} & \gain{45.91}{+13.10}
& \gain{67.55}{+23.95} \\
\quad + GDPO
& \topv{\gain{99.81}{+55.14}} & \gain{82.80}{+1.02} & \gain{82.80}{+46.07} & \gain{88.47}{+34.08}
& \gain{85.94}{+37.95} & \gain{29.91}{-4.02} & \gain{26.79}{+10.27} & \gain{47.55}{+14.74}
& \gain{68.01}{+24.41} \\
\quad + \prismo (ours)
& \gain{98.97}{+54.30} & \topv{\gain{84.11}{+2.33}} & \topv{\gain{84.11}{+47.38}} & \gain{89.06}{+34.68}
& \topv{\gain{88.17}{+40.18}} & \topv{\gain{31.92}{-2.01}} & \topv{\gain{28.57}{+12.05}} & \topv{\gain{49.55}{+16.74}}
& \topv{\gain{69.31}{+25.71}} \\
\bottomrule
\end{tabular}%
}

\vspace{0.75em}
\caption{
\textbf{Main results on BFCL v3 benchmarks}: tool calling results on Qwen2.5-3B-Instruct across all evaluated methods. Acc/R: Percentage of correct answers on RLLA style judgement. Acc/B: Percentage of correct answers on BFCL style judgement. Details of RLLA and BFCL judgement refer to Appendix \ref{sec:bfcl_mec}
}
\vspace{-1em}
\label{tab:bfcl_results}
\resizebox{\ifdim\width>\textwidth\textwidth\else\width\fi}{!}{%
\begin{tabular}{l*{12}{c}}
\toprule
\multicolumn{13}{c}{\textbf{Results on BFCL v3}} \\
\midrule
\multirow{2}{*}{Model / Method}
& \multicolumn{3}{c}{Non-live}
& \multicolumn{3}{c}{Live}
& \multicolumn{3}{c}{Multi-turn}
& \multicolumn{3}{c}{Overall} \\
\cmidrule(lr){2-4} \cmidrule(lr){5-7} \cmidrule(lr){8-10} \cmidrule(lr){11-13}
& Fmt & Acc/R & Acc/B
& Fmt & Acc/R & Acc/B
& Fmt & Acc/R & Acc/B
& Fmt & Acc/R & Acc/B \\
\midrule
\base{Qwen2.5-3B-Instruct}
& \base{13.30} & \base{8.70} & \base{72.17}
& \base{14.06} & \base{5.63} & \base{58.30}
& \base{20.56} & \base{7.28} & \base{17.34}
& \base{15.98} & \base{7.20} & \base{49.27} \\
\quad + GRPO Sum
& 65.74 & 40.87 & 74.78
& 40.87 & 31.16 & \topv{68.02}
& 59.53 & 17.43 & \topv{18.28}
& 64.43 & 27.82 & 51.27 \\
\quad + GRPO Product
& 90.17 & 49.48 & 74.96
& 96.97 & 31.16 & 60.07
& 77.51 & 20.50 & 17.99
& 88.22 & 33.71 & 51.01 \\
\quad + GDPO
& 98.87 & 55.30 & 76.09
& \topv{99.33} & \topv{33.16} & 62.37
& 83.22 & 20.68 & 17.93
& 93.81 & \topv{36.38} & 52.13 \\
\quad + \prismo (ours)
& \topv{99.83} & \topv{56.26} & \topv{80.35}
& \topv{99.33} & 31.09 & 62.22
& \topv{86.54} & \topv{21.24} & 17.81
& \topv{95.23} & 36.20 & \topv{53.46} \\
\bottomrule
\end{tabular}%
}

\vspace{0.75em}
\caption{
\textbf{Main results on Helpfulness-Safety Alignment benchmarks}: helpfulness and safety alignment results on Qwen2.5-3B-Instruct across all evaluated optimization methods.
}
\label{tab:helpfulness_safety_alignment}
\vspace{-1em}
\resizebox{\ifdim\width>\textwidth\textwidth\else\width\fi}{!}{%
\begin{tabular}{l*{9}{c}}
\toprule
\multicolumn{10}{c}{\textbf{Results on Helpfulness-Safety Alignment}} \\
\midrule
\multirow{2}{*}{Model / Method}
& \multicolumn{3}{c}{Alpaca}
& \multicolumn{3}{c}{HH-RLHF}
& \multicolumn{3}{c}{PKU-SafeRLHF} \\
\cmidrule(lr){2-4} \cmidrule(lr){5-7} \cmidrule(lr){8-10}
& Useful & Harmless & Avg
& Useful & Harmless & Avg
& Useful & Harmless & Avg \\
\midrule

\quad + GRPO Sum
& 3.03 & 3.49 & 3.26
& 3.03 & 4.03 & 3.53
& 4.79 & 6.34 & 5.57 \\

\quad + GRPO Product
& 3.02 & 3.48 & 3.25
& 3.02 & 4.03 & 3.53
& 4.79 & 6.34 & 5.57 \\

\quad + GDPO
& 2.95 & 3.44 & 3.20
& 3.03 & 4.02 & 3.53
& 4.79 & 6.33 & 5.56 \\

\quad + \prismo (ours)
& \topv{3.17} & \topv{3.65} & \topv{3.41}
& \topv{3.14} & \topv{4.15} & \topv{3.65}
& \topv{4.86} & \topv{6.36} & \topv{5.61} \\

\bottomrule
\end{tabular}%
}

\end{table*}

\begin{figure*}[htbp]
    \centering
    \includegraphics[width=\linewidth]{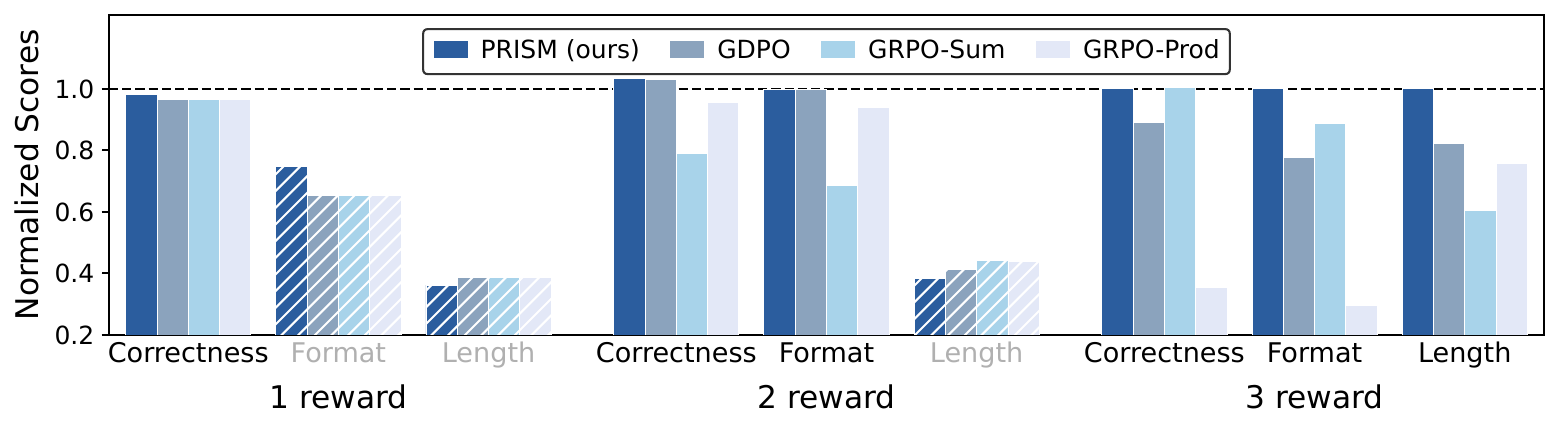}
    \vspace{-1em}
    \caption{Overall BFCL v3 performance under different numbers of training rewards. 1 reward: Train on the \textbf{correctness} reward; 2 reward Trained on \textbf{correctness} and \textbf{format} rewards; 3 reward: Trained on \textbf{correctness}, \textbf{format}, and \textbf{length} rewards. Correctness denotes the RLLA-style accuracy. All scores are normalized with three-reward score of \prismo.}
    \label{fig:reward_scaling_eval}
\end{figure*}

\begin{figure*}[htbp]
    \centering
    \vspace{-1em}
    \includegraphics[width=\linewidth]{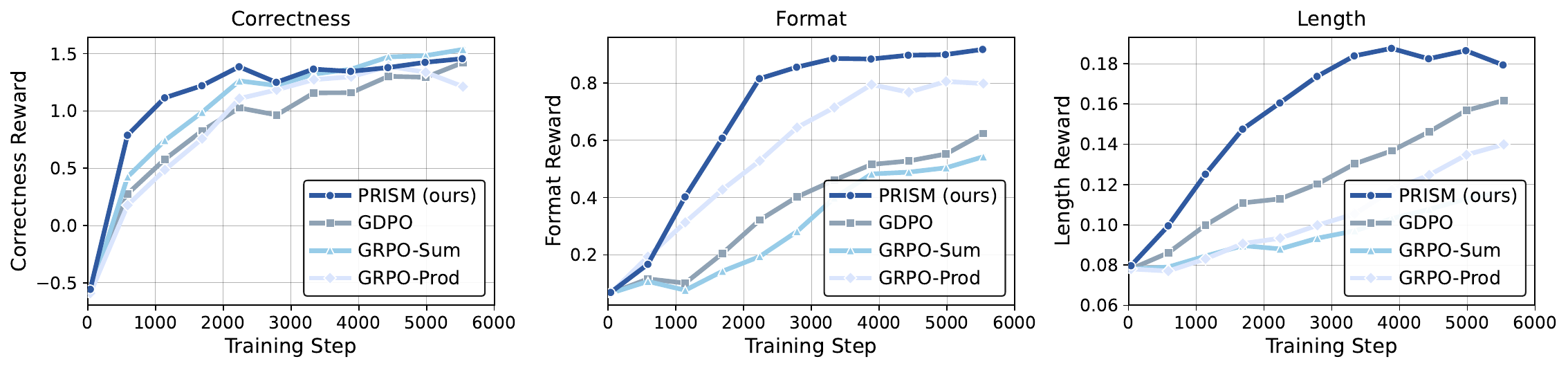}
    \vspace{-2em}
    \caption{Training curves across all methods under the three-reward setting.}
    \vspace{-1em}
    \label{fig:reward_scaling_train}
\end{figure*}

\subsection{Experiment Setup}

We evaluate \prismo on three multi-reward alignment scenarios: scientific question answering, tool-use reasoning, and helpfulness--safety alignment.

\paragraph{Scientific question answering.} For scientific reasoning, we train models on SciKnowEval~\citep{feng2024sciknoweval} and evaluate them on GPQA~\citep{rein2023gpqa} and ScienceQA~\citep{saikh2022scienceqa}. Further details are in Appendix~\ref{sec:sqa_details}.

\paragraph{Tool-use reasoning.} For tool-use reasoning, we train models on the dataset collected by ToolRL~\citep{qian2026toolrl} and evaluate them on BFCL-v3~\citep{patil2025bfcl}, a benchmark evaluates the LLM's ability to call functions. Further details are in Appendix~\ref{sec:bfcl_eval}.

\paragraph{Helpfulness-safety alignment.} For helpfulness-safety alignment, we use the Alpaca dataset~\citep{taori2023alpaca} for policy optimization and evaluate the trained models on the corresponding held-out Alpaca test split. We further evaluate models on the HH-RLHF~\citep{ganguli2022red} and PKU-SafeRLHF~\citep{dai2024safe} datasets. Further details are in Appendix~\ref{sec:hsa_detail}.

\paragraph{Baselines.} We conduct experiments on DeepSeek-R1-1.5B~\citep{shao2024deepseekmath}, Qwen2.5-1.5B-Instruct and Qwen2.5-3B-Instruct~\citep{qwen2.5}. For each backbone and task, we compare \prismo with representative multi-reward RL baselines, including GRPO~\citep{shao2024deepseekmath} variants and GDPO~\citep{liu2026gdpogrouprewarddecouplednormalization}. The GRPO variants include reward weighted summation, denoted as GRPO Sum, and reward product aggregation, denoted as GRPO Prod. Detailed training configurations and hyperparameters are provided in Appendix~\ref{app:th-sqa}, \ref{app:th-tc} and \ref{app:th-hsa}.

\subsection{Main Results}

Tables \ref{tab:main_results}, \ref{tab:bfcl_results}, and \ref{tab:helpfulness_safety_alignment} summarize the main evaluation results across scientific reasoning, tool use, and helpfulness-safety alignment.

On scientific reasoning (Table~\ref{tab:main_results}), \prismo attains the best overall score on all three backbones: 62.73 vs.\ 44.96 for the strongest baseline on DeepSeek-R1-1.5B ($+$17.77), 71.34 vs.\ 63.30 on Qwen2.5-1.5B-Instruct ($+$8.04), and 69.31 vs.\ 68.71 on Qwen2.5-3B-Instruct. The advantage is most pronounced on the harder GPQA benchmark, where \prismo reaches 47.55 average on DeepSeek-R1-1.5B, more than doubling the best baseline (22.69). Crucially, \prismo achieves the highest Joint score in all six benchmark--backbone settings (e.g., 68.51 vs.\ 58.04 on ScienceQA and 28.35 vs.\ 9.60 on GPQA with DeepSeek-R1-1.5B), confirming that it best improves the ability to satisfy all reward criteria \emph{simultaneously}.

On tool calling (Table~\ref{tab:bfcl_results}), \prismo obtains the best overall Fmt (95.23 vs.\ 93.81 for GDPO) and Acc/B (53.46 vs.\ 52.13), while its overall Acc/R (36.20) is on par with GDPO (36.38). It sweeps all three metrics in the Non-live setting (99.83 Fmt, 56.26 Acc/R, 80.35 Acc/B) and achieves the best Fmt and Acc/R in Multi-turn, delivering the most balanced profile across settings.

On helpfulness--safety alignment (Table~\ref{tab:helpfulness_safety_alignment}), \prismo scores highest on \emph{both} usefulness and harmlessness across all three evaluation sets, lifting the average to 3.41 on Alpaca ($+$0.15 over the best baseline), 3.65 on HH-RLHF ($+$0.12), and 5.61 on PKU-SafeRLHF ($+$0.04), whereas the baselines are nearly indistinguishable from one another. These consistent gains under continuous reward-model feedback show that policy-space decomposition remains effective beyond rule-based reward settings.
Compared with all other multi-reward RL baselines, \prismo demonstrates superior or more stable performance in most settings, further validating its effectiveness in multi-objective optimization scenarios.

\subsection{Alignment Tax under Varying Rewards}

Based on the BFCL-v3~\citep{patil2025bfcl} experiments, we further analyze the impact of varying the number of rewards on different algorithms. Specifically, we train the models with one, two and three rewards, respectively.

All algorithms are trained for 6,000 steps, and the final model performance is recorded. To ensure comparability across different metrics, all metrics are normalized using the corresponding rewards of \prismo under three-reward setting.

As shown in Figure~\ref{fig:reward_scaling_eval}, \prismo maintains nearly the same correctness performance after adding format and length rewards, while consistently outperforming all baselines. In contrast, GDPO \citep{liu2026gdpogrouprewarddecouplednormalization} and GRPO-Prod exhibit a clear drop in correctness as the number of rewards increases; although GRPO-Sum maintains relatively stable correctness, its optimization of format and length rewards is substantially weaker than \prismo. These results suggest that \prismo better mitigates the alignment tax and attains a superior Pareto frontier in multi-reward optimization.

Shown in Figure~\ref{fig:reward_scaling_train}, we further visualize the evolution of each reward over training steps under the three-reward setting. Compared with the baselines, \prismo demonstrates substantially higher sample efficiency: it converges to near-optimal performance after roughly 3k steps, while the other methods require more than 6k steps to approach convergence.

\subsection{Inference-Time Controllability}

\begin{figure}[t]
    \centering
    \includegraphics[width=0.6\linewidth]{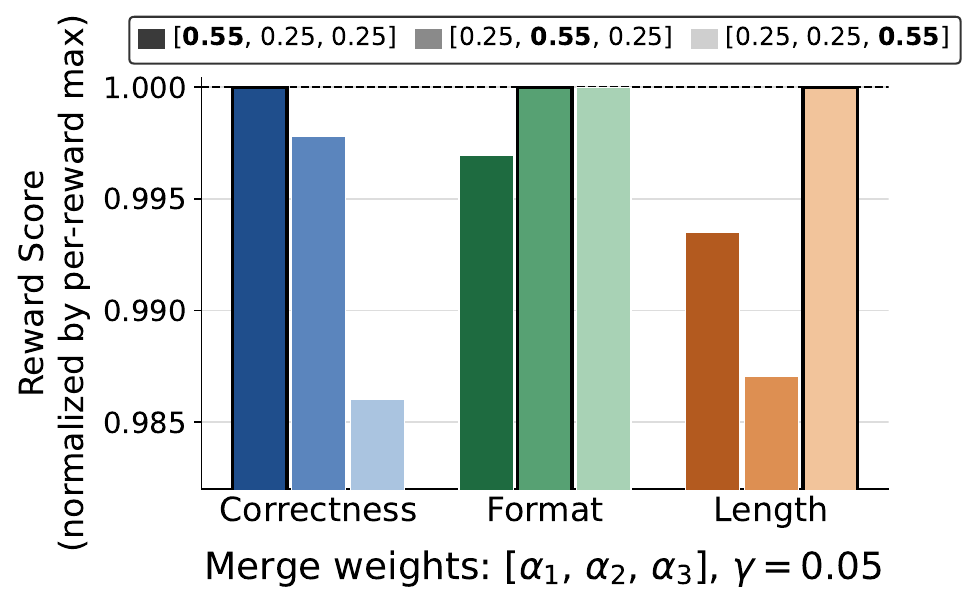}
    \vspace{-1em}
    \caption{Inference-time controllability under different weight configurations on BFCL-Live. Each configuration assigns a dominant weight of $0.55$ to one positive branch while setting the remaining two positive weights to $0.25$.
The negative branch weight is fixed to $\gamma = 0.05$ throughout all experiments.}
    \label{fig:controllable_heatmap}
\end{figure}

To investigate inference-time controllability, we evaluate three weight configurations on the BFCL-Live split. Figure~\ref{fig:controllable_heatmap} shows that increasing the weight $\alpha_k$ consistently improves the corresponding evaluation metric.
Specifically, setting $\alpha_1 = 0.55$ achieves the highest correctness accuracy ($31.53\%$), 
$\alpha_2 = 0.55$ yields the best format accuracy ($98.08\%$), 
and $\alpha_3 = 0.55$ produces the longest average reasoning length ($108.0$). 

These results suggest that the merge weights provide a simple and effective mechanism for controllable inference-time behavior.

\subsection{Ablation Studies}

\paragraph{Ablations on Policy Composition.}
We conduct component ablations on BFCL-v3 using Qwen2.5-3B-Instruct under the same training and evaluation configuration as the main tool-use experiments. We evaluate three variants: (1) replacing reward-specific positive policies with a shared positive policy, (2) removing the global negative policy, and (3) generating rollouts independently from each branch rather than from the composed policy.













\begin{table*}[t]
\centering

\begin{minipage}[t]{0.49\textwidth}
\centering
\small
\setlength{\tabcolsep}{3pt}
\renewcommand{\arraystretch}{1.1}
\captionof{table}{\textbf{Policy-composition ablations} on BFCL-v3 using Qwen2.5-3B-Instruct.}
\label{tab:bfcl_ablation}

\begin{tabular}{lccc}
\toprule
\textbf{Method} & \textbf{Fmt} & \textbf{Acc/R} & \textbf{Acc/B} \\
\midrule
\prismo
& \textbf{95.23} & \textbf{36.20} & \textbf{53.46} \\
\quad w/ shared positive policy
& 94.03 & 34.38 & 52.56 \\
\quad w/o global negative policy
& 94.86 & 34.75 & 52.84 \\
\quad w/ individual-branch rollouts
& 94.84 & 34.59 & 52.48 \\
\bottomrule
\end{tabular}
\end{minipage}
\hfill
\begin{minipage}[t]{0.49\textwidth}
\centering
\small
\setlength{\tabcolsep}{3pt}
\renewcommand{\arraystretch}{1.1}
\captionof{table}{\textbf{Negative-policy weighting ablations} on BFCL-v3 using Qwen2.5-3B-Instruct.}
\label{tab:negative-policy-weighting}

\begin{tabular}{lccc}
\toprule
\textbf{Method} & \textbf{Fmt} & \textbf{Acc/R} & \textbf{Acc/B} \\
\midrule
\prismo
& \textbf{95.23} & \textbf{36.20} & \textbf{53.46} \\
\quad w/ $\max$ weighting
& 95.02 & 33.70 & 52.43 \\
\quad w/ $\operatorname{LogAvgExp}$ weighting
& 95.06 & 34.58 & 52.62 \\
\quad w/ $\operatorname{mean}$ weighting
& 93.27 & 33.90 & 53.12 \\
\bottomrule
\end{tabular}
\end{minipage}

\end{table*}

Table~\ref{tab:bfcl_ablation} summarizes the overall results. All three ablations degrade overall performance. Sharing the positive policy causes the largest drops in format and RLLA accuracy, confirming the importance of preserving reward-specific optimization directions. Removing the global negative policy reduces all overall metrics, supporting its role in modeling shared failure modes. Individual-branch rollouts obtain the lowest overall BFCL accuracy, indicating that sampling from the composed policy better aligns the training distribution with composed decoding. 

\paragraph{Ablations on Negative Weighting Function.}
As Table~\ref{tab:negative-policy-weighting}, we further study the weighting function used to train the global
negative policy. For the $i$-th sampled response, let
$\{A_i^k\}_{k=1}^{N}$ denote its reward-specific advantages.
We compare three alternative weighting functions. The $\max$
variant uses $w_i^-=-\max_{k=1,\dots,N} A_i^k$; the
$\operatorname{LogAvgExp}$ variant uses
$w_i^-=-\exp\!\left(\frac{1}{N}\sum_{k=1}^{N}\log(A_i^k)\right)$;
and the $\operatorname{mean}$ variant uses
$w_i^-=-\frac{1}{N}\sum_{k=1}^{N}A_i^k$.

\section{Related Work}

\paragraph{Reinforcement Learning for LLM.}
RL is the dominant tool for aligning LLMs with human preferences. RLHF-PPO \citep{ouyang2022training, schulman2017proximal} optimizes a learned reward model with a separate value network, while DPO \citep{rafailov2023direct} recasts preference optimization as a stable supervised-style objective. Recent critic-free variants improve scalability and stability: GRPO \citep{shao2024deepseekmath} estimates advantages via group-wise normalization, GSPO \citep{zheng2025group} reduces importance-ratio variance using sequence-level ratios, and DAPO \citep{yu2026dapo} refines the training recipe at scale. All of them, however, optimize a single scalar reward, so multiple preference signals must be collapsed before training.

\paragraph{Multi-Reward RL.}
Since alignment must satisfy correctness, safety, helpfulness, and format compliance at once, most multi-reward methods scalarize rewards into one weighted objective \citep{zhou2024beyond, williams2024multi, quan2024dmoerm}, which is sensitive to reward scales and weights and induces interference among heterogeneous preferences. Refinements include fine-grained rewards \citep{wu2023fine}, adaptive weighting \citep{de2024dynamic}, reward normalization \citep{liu2026gdpogrouprewarddecouplednormalization}, constrained optimization \citep{achiam2017constrained}, model merging \citep{jang2023personalized}, and preference conditioning \citep{yang2024rewards}, yet all of them compose preferences before the policy update. In contrast, \prismo keeps rewards as separate optimization directions and composes the resulting policies in logit space, shifting preference composition from reward space to policy space.

\section{Conclusion}

In this work, we propose \prismo, a policy-space composition framework for multi-reward LLM alignment. Instead of scalarizing heterogeneous rewards into a single optimization signal, \prismo keeps each reward as a separate optimization direction by learning one positive policy per reward. A single global negative policy further captures the union of failure modes. These policies are composed through explicit merge weights, enabling controllable preference trade-offs without retraining. Experiments on scientific reasoning, tool-use reasoning, and helpfulness--safety alignment show that \prismo consistently outperforms reward-space baselines, remains more robust as reward complexity increases, and supports effective inference-time preference control. These results suggest that mixing policies rather than rewards offers a more effective and controllable path toward multi-reward aligned language models.

\bibliographystyle{plainnat}
\bibliography{custom}

\newpage
\appendix

\section{Experimental details of scientific question answering}
\label{sec:detail_sqa}

\subsection{Dataset and Prompt Format}
\label{sec:detail_sqa_format}
We train on the SciKnowEval dataset, restricting to the multiple-choice subset. Each example is formatted as a two-turn chat prompt.

The system prompt is:

\begin{conversationbox}[System Prompt]
You are a helpful AI assistant specialized in solving multiple-choice science questions.
 
For every request, solve the problem carefully and step by step before giving the final answer.
 
You must respond in exactly the following format:
 
<think>
Provide a clear, step-by-step solution.
</think>
<answer>
Provide only the final answer.
</answer>
 
Rules:

1. The <think> section must contain the complete reasoning process.

2. The <answer> section must appear after </think>.

3. The <answer> section must contain only one option letter: A, B, C, or D.

4. Do not include explanation, extra words, or additional formatting in <answer>.

5. Always complete the reasoning before writing the final answer.

\end{conversationbox}

The user turn is rendered as:

\begin{conversationbox}[User Prompt]
Given a question and four options, please select the right answer. Your answer should be "A", "B", "C" or "D".

\{question\}

Choices:

A. \{choice\_a\}

B. \{choice\_b\}

C. \{choice\_c\}

D. \{choice\_d\}

\end{conversationbox}

\subsection{Reward Functions}

Training uses two reward signals: correctness and format.

\paragraph{Correctness Reward.}
The correctness reward $R_{\mathrm{correct}} \in \{0,1\}$ evaluates whether the predicted answer matches the reference answer:
\[
R_{\mathrm{correct}} =
\mathbf{1}[a(x)=y],
\]
where $y$ denotes the reference option letter, and $a(x)$ is the option extracted from the model completion $x$.

\paragraph{Format Reward.}
The format reward $R_{\mathrm{format}} \in \{0,1\}$ is defined as
\[
R_{\mathrm{format}}
=
\mathbf{1}[x \text{ matches the target format}],
\]
{\raggedright
where the target format is
\texttt{<think>\ldots</think><answer>\ldots</answer>}.\par}

\subsection{Training Hyperparameters}
\label{app:th-sqa}
We train Qwen2.5-1.5B-Instruct and Qwen2.5-3B-Instruct for 1 epoch, and DeepSeek-R1-Distill-Qwen-1.5B for 1.5 epochs. Other training hyperparameters are summarized in Table~\ref{tab:sciknoweval-shared-config}.
\begin{table}[!htbp]
\centering
\small
\caption{Scientific question answering training configuration.}
\label{tab:sciknoweval-shared-config}

\begin{tabular}{lc}
\toprule
Setting & Value \\
\midrule

LoRA target modules &
\begin{tabular}[c]{@{}c@{}}
\texttt{q,k,v,o}\\
\texttt{gate,up,down\_proj}
\end{tabular} \\

LoRA rank / alpha / dropout
& 32 / 64 / 0.05 \\

Batch size
& 512 \\

Group size
& 4 \\

Max prompt length
& 1024 \\

Max completion length
& 1000 \\

Learning rate
& $1\times10^{-5}$ \\

\bottomrule
\end{tabular}
\end{table}

\subsection{Evaluation on ScienceQA and GPQA}
\label{sec:sqa_details}

Each model is required to first produce its reasoning process enclosed by \texttt{<think></think>}, followed by a final answer enclosed by \texttt{<answer></answer>}. The order of these two fields must be preserved. We use two rewards in this setting. The format reward verifies whether the output contains both \texttt{<think></think>} and \texttt{<answer></answer>} in the correct order. The correctness reward evaluates whether the answer extracted from the \texttt{<answer></answer>} field matches the ground-truth answer.

We report pass@1 accuracy under greedy decoding, together with format adherence rate.
All evaluation samples follow the same prompt format and answer-extraction protocol as described in Section~\ref{sec:detail_sqa_format}.

For ScienceQA, we use text-only samples from the natural science and closed-choice subset.
For GPQA, we use the \texttt{gpqa\_main} split and render each sample as a four-choice question, with answer options deterministically shuffled per sample.

\section{Experimental Details of Tool-use Reasoning}
\subsection{Training Prompt Format}
\label{sec:detail_toolrl_format}

\begin{conversationbox}[Prompt Format for ToolRL Training]
You are a helpful dialogue assistant capable of leveraging tool calls to solve user tasks and provide structured chat responses.

\textbf{Available Tools}

In your response, you can use the following tools:

\{\{ Tool List \}\}

\textbf{Steps for Each Turn}

1. \textbf{Think:} Recall relevant context and analyze the current user goal.

2. \textbf{Decide on Tool Usage:} If a tool is needed, specify the tool and its parameters.

3. \textbf{Respond Appropriately:} If a response is needed, generate one while maintaining consistency across user queries.

\textbf{Output Format}

\texttt{<think>} Your thoughts and reasoning \texttt{</think>}

\texttt{<tool\_call>}
\{"name": "Tool name", "parameters": \{"Parameter name": "Parameter content", "...": "..."\}\}

\{"name": "...", "parameters": \{"...": "...", "...": "..."\}\}

...

\texttt{</tool\_call>}

\texttt{<response>} AI's final response \texttt{</response>}

\textbf{Important Notes}

1. You must always include the \texttt{<think>} field to outline your reasoning. Provide at least one of \texttt{<tool\_call>} or \texttt{<response>}. Decide whether to use \texttt{<tool\_call>} possibly multiple times, \texttt{<response>}, or both.

2. You can invoke multiple tool calls simultaneously in the \texttt{<tool\_call>} fields. Each tool call should be a JSON object with a \texttt{"name"} field and a \texttt{"parameters"} field containing a dictionary of parameters. If no parameters are needed, leave the \texttt{"parameters"} field as an empty dictionary.

3. Refer to the previous dialogue records in the history, including the user's queries, previous \texttt{<tool\_call>}, \texttt{<response>}, and any tool feedback noted as \texttt{<obs>} if exists.
\end{conversationbox}

\begin{conversationbox}[User Prompt for ToolRL Training]
\textbf{Dialogue History}

\texttt{<user>} \{\{ Initial User Input \}\} \texttt{</user>}

\texttt{<think>} Round 1 Model Thought \texttt{</think>}

\{\{ Round 1 model output \texttt{<tool\_call>} or \texttt{<response>} \}\}

\texttt{<obs>} Round 1 Observation \texttt{</obs>}

...

\texttt{<user>} \{\{ User Input \}\} \texttt{</user>}

...
\end{conversationbox}

\subsection{Tool Calling Reward Functions}

\paragraph{Format Reward.}
The format reward $R_{\mathrm{format}} \in \{0,1\}$ evaluates whether a model output follows the tag structure required by the ground-truth answer. The output must begin with a \texttt{<think>} section and then contain the required \texttt{<tool\_call>} and/or \texttt{<response>} sections in the prescribed order. When a tool-call or response section is required, its opening and closing tags must each appear exactly once.

\paragraph{Correctness Reward.}
The correctness reward $R_{\mathrm{correct}} \in [-3,3]$ quantifies how well the predicted tool calls match the ground-truth tool calls. The comparison includes:

\begin{enumerate}
\item \textbf{Tool name matching}: a multiset overlap score in $[0,1]$ is computed between the predicted and ground-truth tool names;
\item \textbf{Parameter matching}: ground-truth calls are processed one by one. For each ground-truth call, every unmatched predicted call with the same tool name is assigned a parameter-level score. This score is the sum of a parameter-name overlap term in $[0,1]$ and one point for each ground-truth parameter whose value is predicted exactly. The highest parameter-level score is added to the total score, and the corresponding predicted call is marked as matched. Each predicted call can therefore be matched at most once.
\end{enumerate}

The tool-name score and all selected parameter-level scores are summed and divided by the maximum attainable score for the ground-truth calls. The resulting normalized score in $[0,1]$ is then linearly mapped to $[-3,3]$.

\paragraph{Length Reward.}
The length reward $R_{\mathrm{length}} \in [0,1]$ encourages the model to produce sufficiently detailed reasoning within the \texttt{<think>} section. It increases linearly with the length of the reasoning content up to the predefined threshold of 512 words and is capped at $1$ once the threshold is reached. Outputs without a valid \texttt{<think>} section receive a reward of $0$.

\subsection{Training Hyperparameters}
\label{app:th-tc}
Detail training hyperparameters of Tool-Use Reasoning are summarized in Table~\ref{tab:tooluse_multidipole_config}.
\begin{table}[H]
\centering
\small
\setlength{\tabcolsep}{4pt}

\caption{Tool-use reasoning training configuration.}
\label{tab:tooluse_multidipole_config}

\begin{tabular}{lc}
\toprule
Setting & Value \\
\midrule

LoRA target modules &
\begin{tabular}[c]{@{}c@{}}
\texttt{q,k,v,o}\\
\texttt{gate,up,down\_proj}
\end{tabular} \\

LoRA rank / alpha / dropout
& 32 / 64 / 0.05 \\

Batch size
& 2048 \\

Mini batch size
& 128 \\

Group size
& 4 \\

Max prompt length
& 2048 \\

Max completion length
& 1024 \\

Learning rate
& $1\times10^{-5}$ \\

Epochs
& 2 \\

\bottomrule
\end{tabular}
\end{table}

\subsection{BFCL Evaluation}

\subsubsection{Evaluation Protocol}
\label{sec:bfcl_eval}

To evaluate tool-use reasoning on BFCL, we convert BFCL examples into the RLLA protocol instead of using the native BFCL function-calling interface. For each example, we construct an RLLA-style prompt following the same prompt format described in \S\ref{sec:detail_toolrl_format}, including the available tool definitions and dialogue history.

Each model is required to first generate its reasoning process enclosed by \texttt{<think></think>}, and then decide whether to perform a tool call or directly respond to the user. If the model chooses to call a tool, it must output a JSON-formatted tool request enclosed by \texttt{<tool\_call></tool\_call>}. If the model chooses to answer directly, it must output the final response enclosed by \texttt{<response></response>}. We again use two rewards. The format reward checks whether the output follows the required structure, namely \texttt{<think></think>} followed by either \texttt{<tool\_call></tool\_call>} or \texttt{<response></response>}. The correctness reward evaluates whether the answer in \texttt{<response></response>} matches the ground truth, or whether the generated tool call contains the correct fields.

\subsubsection{Evaluation Metrics}
\label{sec:bfcl_mec}

Let $N$ denote the number of evaluated examples. We report both strict RLLA metrics and BFCL-style semantic metrics.

\paragraph{RLLA Accuracy.}
The RLLA accuracy is a strict task-level metric. For examples with reference tool calls, it requires the correctness reward to reach its maximum possible score. For examples without reference tool calls, it requires the model to predict no tool call:

\begin{equation}
\begin{aligned}
\mathrm{rlla\_acc}
=
\frac{1}{N}
\sum_{i=1}^{N}
\mathbf{1}\!\Big[
(G_i \neq \emptyset
\land
R_{\mathrm{correct}}(P_i,G_i)=3)
\lor
(G_i=\emptyset
\land
P_i=\emptyset)
\Big].
\end{aligned}
\end{equation}

\paragraph{BFCL Accuracy.}
The BFCL accuracy evaluates the semantic correctness of predicted tool calls while ignoring RLLA wrapper-format errors. Predicted \texttt{<tool\_call>} blocks are extracted, converted back into BFCL-style function calls, and evaluated using the BFCL checker:

\begin{equation}
\begin{aligned}
\mathrm{bfcl\_acc}
=
\frac{1}{N}
\sum_{i=1}^{N}
\mathbf{1}\!\Big[
\mathrm{BFCLChecker}(P_i,G_i)
=
\mathrm{correct}
\Big].
\end{aligned}
\end{equation}

\paragraph{Format Accuracy.}
The format accuracy measures protocol adherence independently of tool correctness:
\[
\mathrm{format\_acc}
=
\frac{1}{N}
\sum_{i=1}^{N}
\mathbf{1}
\left[
R_{\mathrm{format}}(x_i)=1
\right],
\]
where $R_{\mathrm{format}}$ follows the same definition as the format reward in the training objective.

\paragraph{Length Reward.}
The evaluation-time length reward measures the mean normalized reasoning length using the same definition as the training-time length reward:
\[
\mathrm{length\_reward}
=
\frac{1}{N}
\sum_{i=1}^{N}
\min
\left(
1,
\frac{n_{\mathrm{think}}(x_i)}{L}
\right),
\]
where the maximum reasoning length is
\[
L=512.
\]

\section{Experimental details of helpfulness-safety alignment}
\label{sec:detail_hsa}

\subsection{Training Hyperparameters}
\label{app:th-hsa}
For helpfulness-safety alignment, we train Qwen2.5-3B-Instruct with LoRA and evaluate the checkpoint at step 540. Other training hyperparameters are summarized in Table~\ref{tab:safe-alignment-shared-config}.

\begin{table}[!htbp]
\centering
\small
\caption{Helpfulness-safety alignment training configuration.}
\label{tab:safe-alignment-shared-config}

\begin{tabular}{lc}
\toprule
Setting & Value \\
\midrule

Base model
& \texttt{Qwen2.5-3B-Instruct} \\

LoRA target modules &
\begin{tabular}[c]{@{}c@{}}
\texttt{q,k,v,o}\\
\texttt{gate,up,down\_proj}
\end{tabular} \\

LoRA rank / alpha / dropout
& 32 / 64 / 0.05 \\

Batch size
& 128 \\

Group size
& 4 \\

Max prompt length
& 512 \\

Max completion length
& 1024 \\

Learning rate
& $1\times10^{-6}$ \\

KL coefficient
& 0.001 \\

Training precision
& bfloat16 \\

Training checkpoint
& step 540 \\

\bottomrule
\end{tabular}
\end{table}

\subsection{Evaluation Details of Helpfulness-Safety Alignment}
\label{sec:hsa_detail}
Each model is given a user prompt and is required to produce a response that is both helpful to the user and safe with respect to potentially harmful requests. We use two rewards in this setting. The helpfulness reward is computed using the helpfulness reward model provided by the Amo project~\citep{amo_project}. The safety reward is derived from the corresponding harmlessness cost model from the same project. The overall alignment objective encourages responses that are simultaneously helpful and safe.

\section{Prefix Token Initialization.}

We initialize each learnable prefix embedding as
\[
\begin{aligned}
\mathbf{p}_i^{(0)}
&=
\operatorname{Emb}_{\theta_0}(\texttt{\textbackslash n})
\end{aligned}
\]
where $\operatorname{Emb}_{\theta_0}$ denotes the input embedding layer of the pretrained language model. The newline embedding provides an in-distribution initialization that is compatible with the pretrained model's input embedding space and prompt formatting.

\section{Controllable Inference Cases}

We illustrate inference-time steerability with an example from the BFCL-Live split.

\smallskip
\noindent\textbf{Task.}

\begin{conversationbox}[User]
I'd like to purchase 2 tickets for `After The Wedding'
showing on 2023-04-22 at 19:00 in Berkeley.
\end{conversationbox}

\begin{conversationbox}[Expected Tool Call]
\texttt{Movies\_1\_BuyMovieTickets(\{}
movie\_name,
number\_of\_tickets,
show\_date,
location,
show\_time,
show\_type
\texttt{\})}
\end{conversationbox}

\noindent\textbf{Responses under three weight configurations.}

Weights are ordered as $(\alpha_1,\alpha_2,\alpha_3)$, corresponding to
correctness, format, and length. In each configuration, the dominant branch is
set to $0.55$ and the other two branches are set to $0.25$.

\noindent\textbf{$\alpha_1 =0.55, \; \alpha_2=\alpha_3 = 0.25$ (correctness-dominant).}
The model directly invokes the target purchase tool with all required arguments.

\begin{conversationbox}[Correctness-dominant Inference]
\begin{lstlisting}
<think>
Identifies the movie title, ticket count, date, 
location, and time, then selects the purchase 
tool.  (*@\textbf{  [...length=126 words] }@*)
</think>
<tool_call>
{"name": "Movies_1_BuyMovieTickets",
 "parameters": 
 {"movie_name": "After The Wedding",
                "number_of_tickets": 2,
                "show_date": "2023-04-22",
                "location": "Berkeley, CA",
                "show_time": "19:00",
                "show_type": "regular"}}
</tool_call>
\end{lstlisting}
\end{conversationbox}

\noindent\textbf{$\alpha_2 =0.55, \; \alpha_1=\alpha_3 = 0.25$ (format-dominant).}
The response is syntactically well-formed, but the model calls a show-time lookup tool
rather than completing the purchase.

\begin{conversationbox}[Format-dominant Inference]
\begin{lstlisting}
<think>
Recognizes the ticket-purchase intent, but 
decides to confirm show-time details first. (*@\textbf{  [...length=115 words] }@*)
</think>
<tool_call>
{"name": (*@\texttt{\textcolor{red}{Movies\_1\_GetTimesForMovie}}@*), (*@\textbf{// wrong tool: should call Movies\_1\_BuyMovieTickets}@*)
 "parameters": 
 (*@\texttt{\textcolor{red}{\{}}@*)
 (*@\texttt{\textcolor{red}{"movie\_name": "After The Wedding",}}@*)
 (*@\texttt{\textcolor{red}{"location": "Berkeley, CA",}}@*)
 (*@\texttt{\textcolor{red}{"show\_date": "2023-04-22",}}@*)
 (*@\texttt{\textcolor{red}{"theater\_name": "Any Theater",}}@*)
 (*@\texttt{\textcolor{red}{"show\_type": "regular"}}@*)
 (*@\texttt{\textcolor{red}{\}}}@*) (*@\textbf{// parameters belong to the wrong tool schema}@*)
 }
</tool_call>
\end{lstlisting}
\end{conversationbox}

\noindent\textbf{$\alpha_3 = 0.55, \; \alpha_1=\alpha_2 = 0.25$ (length-dominant).}
The model produces a longer reasoning trace.

\begin{conversationbox}[Length-dominant Inference]
\begin{lstlisting}
<think>
Mentions the movie, date, time, location, and 
ticket count, then gives extra reasoning about 
finding show times before buying tickets. 
(*@\textbf{  [...length=164 words] }@*) (*@\textbf{// longer reasoning trace under the length-dominant inference }@*)
</think>
<tool_call>
{"name": (*@\texttt{\textcolor{red}{Movies\_1\_GetTimesForMovie}}@*), (*@\textbf{// wrong tool: should call Movies\_1\_BuyMovieTickets}@*)
 "parameters": 
 (*@\texttt{\textcolor{red}{\{}}@*)
 (*@\texttt{\textcolor{red}{"movie\_name": "After The Wedding",}}@*)
 (*@\texttt{\textcolor{red}{"location": "Berkeley, CA",}}@*)
 (*@\texttt{\textcolor{red}{"show\_date": "2023-04-22",}}@*)
 (*@\texttt{\textcolor{red}{"show\_type": "regular"}}@*)
 (*@\texttt{\textcolor{red}{\}}}@*) (*@\textbf{// parameters belong to the wrong tool schema}@*)
 }
</tool_call>
\end{lstlisting}
\end{conversationbox}

\section{Inference Efficiency}
\label{sec:inference_efficiency}
We measure the inference cost of \prismo against a single-policy baseline under the same serving configuration on 8 $\times$ NVIDIA H800. We report per-response decoding latency, generation throughput, and peak GPU memory.

\begin{table}[htbp]
\centering
\small
\setlength{\tabcolsep}{4pt}
\renewcommand{\arraystretch}{1.1}

\caption{
\textbf{Inference efficiency} on Qwen2.5-3B-Instruct. Latency is measured in second, throughput in tokens per second, and peak GPU memory in GB. \prismo{} with $N$ rewards decodes $N{+}1$ branches.
}
\label{tab:inference-efficiency}

\begin{tabular}{lccc}
\toprule
Method & Latency & Throughput & Memory \\
\midrule
Single policy       & 2.01 & 127.56 & 63.74 \\
\prismo{} ($N{=}2$) & 2.04 & 125.71 & 63.74 \\
\prismo{} ($N{=}3$) & 2.12 & 121.06 & 63.74 \\
\bottomrule
\end{tabular}
\end{table}

\section{Limitations}

There are several limitations in this study. First, \prismo assumes that each reward can be represented by a distinct positive policy. It also assumes that a single negative policy can capture the union of reward-specific failure modes. This design is effective across all our experimental settings, which cover up to three rewards. Validating it at substantially larger reward counts is left for future work. Second, the current policy-space composition relies on manually specified merge weights, and how to automatically adapt these weights to different prompts, users, or deployment contexts remains underexplored. Third, although per-token latency stays that of a single forward pass, mixture decoding requires multiple prefix-conditioned branches, and its memory and FLOPs scale with the number of rewards. Future work can study more adaptive policy composition mechanisms, more expressive negative-policy designs, and more efficient decoding strategies.

\end{document}